\documentclass{article}
\usepackage{spconf,amsmath,graphicx,amssymb}
\usepackage{multirow}
\usepackage{arydshln}

\title{Unimodal Aggregation For CTC-based Speech Recognition}
\name{Ying Fang$^{1, 2}$, Xiaofei Li$^{2,3}$\sthanks{Corresponding author.}}
\address{
$^{1}$ Zhejiang University, China \\
$^{2}$ School of Engineering, Westlake University, China \\
$^{3}$ Institute of Advanced Technology, Westlake Institute for Advanced Study, China}

\begin{document}
\fontsize{9.5pt}{\baselineskip}\selectfont
\maketitle
\begin{abstract}
This paper works on non-autoregressive automatic speech recognition. A unimodal aggregation (UMA) is proposed to segment and integrate the feature frames that belong to the same text token, and thus to learn better feature representations for text tokens. The frame-wise features and weights are both derived from an encoder. Then, the feature frames with unimodal weights are integrated and further processed by a decoder. Connectionist temporal classification (CTC) loss is applied for training. Compared to the regular CTC, the proposed method learns better feature representations and shortens the sequence length, resulting in lower recognition error and computational complexity. Experiments on three Mandarin datasets show that UMA demonstrates superior or comparable performance to other advanced non-autoregressive methods, such as self-conditioned CTC. Moreover, by integrating self-conditioned CTC into the proposed framework, the performance can be further noticeably improved. 
\end{abstract}
\begin{keywords}
unimodal aggregation, non-autoregressive speech recognition, CTC 
\end{keywords}

\section{Introduction}
\label{sec:intro}
End-to-end automatic speech recognition (ASR) has been largely developed recently.  Encoder-decoder attention mechanism \cite{chan2016listen} and connectionist temporal classification (CTC) \cite{graves2014towards} are two major techniques for end-to-end ASR. One key difficulty for ASR is to automatically align the input acoustic feature frames to the output text tokens. The autoregressive (AR) attention mechanism uses a decoder to aggregate input frames and predict text tokens one by one, where one text token is predicted conditioning on the previously predicted token. 
CTC is a non-autoregressive (NAR) method that uses only an encoder. It allows for independent and parallel prediction for all frames, resulting in much faster inference than autoregressive methods. However, the independence assumption of CTC will lose the dependencies of tokens, leading to reduced recognition performance. 

CTC-based ASR has attracted lots of research attention. Intermediate CTC \cite{lee2021intermediate} regularizes the CTC training by combining the CTC loss of intermediate layers. Self-conditioned CTC \cite{nozaki2021relaxing} embeds the CTC predictions of intermediate layers to the forward data flow to relax the independence assumption of CTC, as the final outputs are conditioned on some intermediate token predictions. 
There are many NAR methods developed in the encoder-decoder framework as well, where the decoder is often deployed to aggregate acoustic information and/or learn the dependencies between text tokens in a NAR way. Mask-CTC \cite{higuchi2020mask} uses the decoder to perform masked token prediction, where the low-confidence CTC predictions are taken as masked tokens. LASO (Listen Attentively, and Spell Once) \cite{bai2021fast} directly predicts all the text tokens in parallel with the decoder. CIF (Continuous Integrate-and-Fire) \cite{dong2020cif} proposes a frame-wise weight
and one text token is detected by accumulating weights until 1. Paraformer \cite{Gao2022ParaformerFA} utilizes the CIF-based predictor and a glancing language model (GLM) sampler to address the issue of token length prediction and interdependence modeling.  \cite{deng2022improving} proposes to integrate the pre-trained large acoustic and language models.
A comparative study is conducted in \cite{higuchi2021comparative} to compare various NAR models.

%

This work proposes a simple yet effective unimodal aggregation (UMA) for NAR ASR. 
The frame-wise aggregation weights are predicted based on the outputs of an encoder. The continuous frames with unimodal weights, namely have first increasing aggregation weights and then decreasing aggregation weights, are considered to belong to the same text token and integrated together, and then further processed by a decoder. After frame integration, the sequence length will be reduced at token level, but it is not forced to equal the length of the token sequence precisely: non-speech frames could be integrated independently into speech frames; speech frames belonging to one text token possibly have multimodal aggregation weights. Therefore, the CTC loss is still used to tackle the non-speech and repeated-speech cases. 
The proposed method explicitly segments and integrates the feature frames, and thus could learn better feature representations for text tokens compared to the regular CTC. 
In addition, the shorter sequence length after integration will reduce the computation complexity. 
Currently, the proposed unimodal aggregation is only suitable for monosyllable languages with clear acoustic boundaries, such as Chinese. 
In addition, we explore the integrated framework with self-conditioned CTC \cite{nozaki2021relaxing} to further improve the UMA accuracy and alleviate the independence assumption of CTC.

\section{Method}
\label{sec:method}

\subsection{CTC Review}
\label{sec:ctc}

In ASR, $\mathbf{x}=(x_1, \dots, x_t, \dots, x_T) \in \mathbb{R}^{D\times T}$ is a speech feature sequence, and $\mathbf{l}=(l_1,\dots,l_u,\dots,l_U) \in \mathbb{K}^{U}$ is the corresponding text token sequence, where $D$ denotes the dimension of the speech feature, $t\in[1,T]$ and $u\in[1,U]$ respectively denote the time index of feature sequence and text token sequence, and normally $T\gg U$. The dictionary/alphabet $\mathbb{K}$ consists of $K$ predefined text tokens, such as phonemes, characters, graphemes, or words. 

To map the input feature sequence to the output token sequence, a new alphabet is defined as $\mathbb{K}'=\mathbb{K}\cup\{\epsilon\}$ in CTC, where $\epsilon$ denotes a blank token. CTC uses a neural network to classify each time step of the speech feature sequence into text tokens. The network output is $\mathbf{y}=\{y^t_k\}_{t=1,k=1}^{T,K+1}$, and $y^t_k$ denotes the probability of classifying $x_t$ to the $k$-th text token. One prediction path $\mathbf{\pi}=(\pi_1,\dots,\pi_t,\dots,\pi_T)\in\mathbb{K}'^T$ is assumed to be conditionally independent, and its conditional probability is $p(\mathbf{\pi}|\mathbf{x})=\prod_{t=1}^T y_{\pi_t}^t$.  Some prediction paths can be mapped to the target token sequence $\mathbf{l}$ by removing first the repeated tokens and then the blank tokens. Let $\Pi$ denote the set of such prediction paths, then the conditional probability of $\mathbf{l}$ can be computed as $p(\mathbf{l}|\mathbf{x}) = \sum_{\mathbf{\pi} \in \Pi} p(\mathbf{\pi}|\mathbf{x})$. CTC trains the neural network to maximize this conditional probability using a forward-backward algorithm. At inference, the ASR result can be obtained with the prediction paths. 

Empirically, the CTC network learns to aggregate information from multiple input frames that belong to the same text token, and outputs spike predictions of text tokens. CTC is powerful as it is able to automatically align a long input sequence to a short output sequence by assigning blank tokens and repeated speech tokens. However, implicitly aggregating and aligning multiple input frames may not be an easy task, and the aggregation error may lead to poor token representation and thus recognition error.  

\begin{figure}[tb]
\centering
\centerline{\includegraphics[width=0.85\columnwidth]{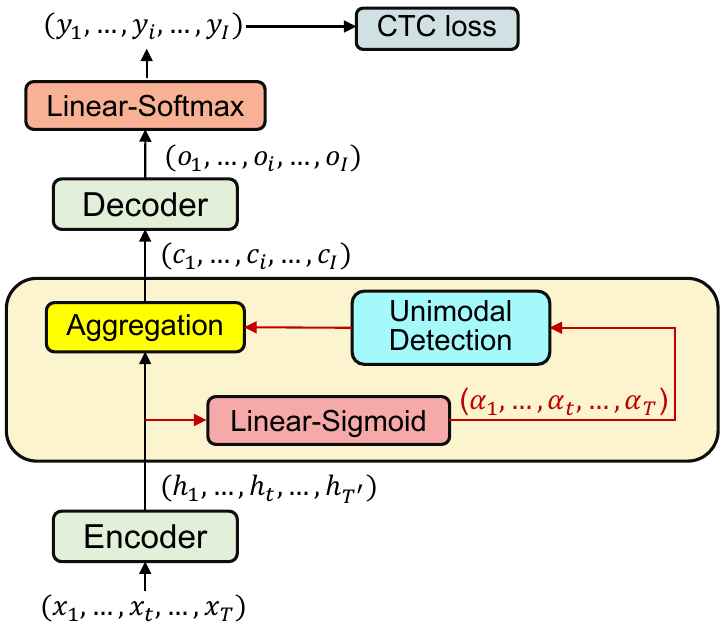}}
\caption{Network architecture of unimodal aggregation.}
\label{fig:nn}
\vspace{-0.4cm}
\end{figure}

\subsection{Proposed Method}
\label{sec:proposed}

This work proposes a simple yet effective unimodal aggregation to explicitly integrate multiple input frames that are likely to belong to the same text token, which will decrease the difficulty of information aggregation for the network. 

The proposed ASR model includes an encoder, an unimodal aggregation, and a decoder. Note that the encoder network here is flexible and can be Transformer, Conformer \cite{gulati2020conformer}, E-Branchformer \cite{kim2023branchformer}, etc. In contrast, the decoder is fixed and has a similar structure to that of the Transformer encoder, namely a NAR self-attention network. 
Fig.~\ref{fig:nn} shows the network architecture of unimodal aggregation. 

The encoder transforms the input feature sequence $\mathbf{x}$ to a hidden feature sequence $\mathbf{h}=(h_1, \dots, h_t, \dots, h_{T'})$, which is down-sampled relative to the input sequence by a factor of 4. Then, a scalar aggregation weight is predicted for each time step with a Linear-Sigmoid network as:
\begin{align}
\alpha_t=\text{Sigmoid}(\text{Linear}(h_t)),
\end{align}
The frames that have unimodal aggregation weights, namely first increasing weights and then decreasing weights, are considered to belong to the same text token. This is based on an intuitive guess and observing the
properties of Transformer attention that a monotonic attention
mechanism should have attention valleys at the boundaries of
text tokens. An example of aggregation weights can be seen in Fig.~\ref{fig:example}.

An aggregation weight valley $t$ is defined as the time step with $\alpha_t\le \alpha_{t-1} \ \text{and} \ \alpha_t \le \alpha_{t+1}$. The time index of aggregation weight valley is denoted as $\tau_i\in[1, T']$, where $i$ denotes the index of the valley. The hidden features are then integrated based on the unimodal aggregation weights:
\begin{align}
c_i=\frac{\sum_{t=\tau_i}^{\tau_{i+1}+1}\alpha_t h_t}{\sum_{t=\tau_i}^{\tau_{i+1}+1}\alpha_t}.
\end{align}
Note that, the frame range for the $i$-th token is set to $t\in[\tau_i,\tau_{i+1}+1]$, which will have two overlap frames with its neighboring tokens, such as the frame range for the $\{i+1\}$-th token is $t\in[\tau_{i+1},\tau_{i+2}+1]$. Setting two overlap frames is very important for training the UMA model successfully. Specifically, the weights of overlapped frames can be adjusted by gradient descent from two tokens, and thus the aggregation weight valleys can flexibly shift during training. 

In addition, to not discard frames at the beginning and end of the sentence, the first ($t=1$) and the end ($t=T'$) time steps are forced to be aggregation weight valleys. 

The integrated hidden feature sequence is denoted as $\mathbf{c} = (c_1,\dots,c_i,\dots,c_I)$. Finally, the decoder transforms $\mathbf{c}$ to an output sequence $\mathbf{o} = (o_1,\dots,o_i,\dots,o_I)$, which is followed by a Linear-Softmax network to predict $\mathbf{y}=\{y_t^k\}_{i=1,k=1}^{I,K+1}$. Importantly, after frame integration, the sequence length $I$ will be at the token level. However, $I$ is not guaranteed to equal the length of the token sequence, as the non-speech frames could be integrated into speech frames; and the frames of one recognition token possibly have multimodal aggregation weights. Therefore, the CTC loss is still used to tackle the non-speech and repeated-speech cases. 

Note that, there is no explicit constraint being put on the aggregation weights $\alpha_t$, and the CTC loss automatically learns to assign unimodal aggregation weights to the feature frames of each token.

\section{Experiments}
\label{sec:experiments}

\subsection{Datasets}
Experiments are conducted on three Mandarin Chinese datasets, i.e. AISHELL-1 \cite{bu2017aishell} (178 hours), AISHELL-2 \cite{du2018aishell} (1000 hours) and HKUST \cite{liu2006hkust} (149 hours). AISHELL-1 was recorded with a high-fidelity microphone. AISHELL-2 was recorded with an iPhone, and the parallel recording with a high-fidelity microphone and an Android phone are also used for development and test. HKUST was recorded through spontaneous conversations during phone calls. 4,232, 5,211, and 3,655 Mandarin characters are taken as recognition tokens in AISHELL-1, AISHELL-2, and HKUST, respectively. 

\begin{table*}[t]
\centering
\vspace{-0.3cm}
\caption{Character error rate (CER,\%) with three different encoders on HKUST.}
\label{tab:hkust}
\setlength{\tabcolsep}{2.5mm}{
\begin{tabular}{cl|cccc|cccc|cccc}
\hline
\multicolumn{2}{c|}{\multirow{2}{*}{\textbf{Model}}} & \multicolumn{4}{c|}{\textbf{Transfomer}} & \multicolumn{4}{c|}{\textbf{Conformer}} & \multicolumn{4}{c}{\textbf{E-Branchformer}} \\ 
\multicolumn{2}{c|}{} & sub & del & ins & CER & sub & del & ins & CER & sub & del & ins & CER \\ 
\hline
\multirow{2}{*}{\rotatebox{90}{AR}} 
& Hybrid CTC/Attention & 18.0 & 2.9 & 3.2 & 24.0 & 16.9 & 3.1 & 3.3 & 23.3 & 15.2 & 2.3 & 3.1 & 20.6 \\
& \ + beam search & 15.9 & 2.8 & 2.8 & \textbf{21.6} & 15.7 & 2.5 & 3.0 & \textbf{21.2} & 14.1 & 2.3 & 2.8 & \textbf{19.3} \\ 
\hdashline
\multirow{4}{*}{\rotatebox{90}{NAR}}
& CTC & 18.4 & 3.0 & 3.3 & 24.7 & 17.3 & 2.8 & 3.2 & 23.2 & 16.0 & 2.6 & 2.9 & 21.6 \\
& Self-conditioned CTC & 18.3 & 2.9 & 3.3 & 24.5 & 16.3 & 2.6 & 3.2 & 22.1 & 14.9 & 2.5 & 3.0 & 20.4 \\
& UMA (prop.) & 15.9 & 6.5 & 2.6 & 25.0 & 15.6 & 2.7 & 3.2 & 21.4 & 14.1 & 3.4 & 2.6 & 20.1 \\
& \ + self-condition & 15.8 & 3.9 & 2.8 & \textbf{22.6} & 14.4 & 2.6 & 3.1 & \textbf{20.0} & 13.7 & 2.6 & 2.9 & \textbf{19.2} \\ 
\hline
\end{tabular}
}
\vspace{-0.4cm}
\end{table*}

\subsection{Model Configuration}
We released our codes on our website\footnote{https://github.com/Audio-WestlakeU/UMA-ASR}. The proposed method is implemented using ESPnet \cite{watanabe2018espnet}.
The encoder is the same as one of the hybrid CTC/attention model \cite{watanabe2017hybrid}, while the decoder blocks are always set as the Transformer encoder blocks. After unimodal aggregation, the integrated feature sequence and a new positional embedding are transformed with a Linear network, and then fed to the decoder. 
The AR hybrid CTC/attention, NAR CTC and self-conditioned CTC \cite{nozaki2021relaxing} are taken as comparison methods. 
The hybrid CTC/attention model is evaluated with greedy decoding or with beam search (beam size is set to 10).
The number of encoder and decoder blocks for hybrid CTC/attention and the proposed model are set to 12 and 6, respectively. CTC and Self-conditioned CTC only use an encoder, thus the number of blocks is set to 18 for a fair comparison. 
On AISHELL-1 and AISHELL-2, we employ the Conformer as the encoder. 
For HKUST, we conduct experiments on three distinct encoder architectures: Transformer, Conformer, and E-Branchformer. A small network is used for AISHELL-1 and HKUST with the number of heads and model dimension being 4 and 256, while a large one is used for AISHELL-2 with the number of heads and model dimension being 8 and 512, respectively.    
We set the feed-forward inner dimension to 2048 for Conformer and Transformer, and to 1024 for E-Branchformer. 
Additionally, we compare with several recently proposed and well-performed NAR methods on AIHSELL-1 and AISHELL-2, including \cite{bai2021fast,dong2020cif,Gao2022ParaformerFA}, and the results are directly quoted from their papers.

\begin{figure}[t]
\centering
\centerline{\includegraphics[width=1\columnwidth]{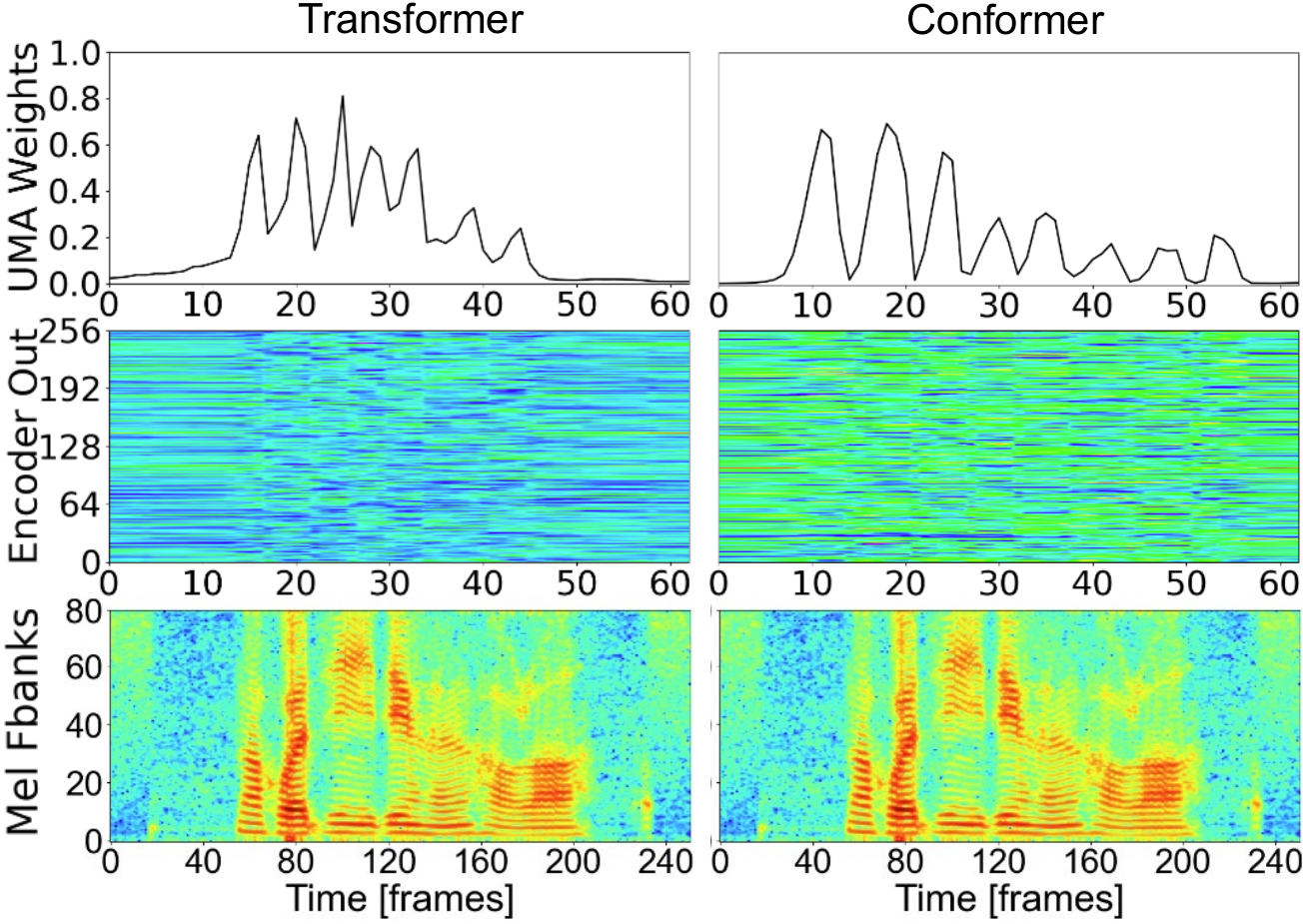}}
\caption{An example of UMA weights for Transformer and Conformer encoder. From bottom to top: Mel spectrogram, output hidden units of the encoder, and UMA weights.}
\label{fig:example}
\vspace{-0.4cm}
\end{figure}

The ESPnet recipes \cite{watanabe2018espnet} for the hybrid CTC/attention are directly used. CTC and self-conditioned CTC are realized by modifying the recipes to have 0 weight for attention loss and changing the number of encoder blocks. There are two exceptions: i) there is no E-Branchformer configuration for HKUST in ESPNet, so we use the E-Branchformer configuration for AISHELL-1 to implement it, which is reasonable as the size of HKUST and AISHELL-1 are similar; ii) similar to the findings in \cite{peng2023comparative}, we observed the training instability when using a large Conformer encoder for CTC and self-conditioned CTC. 
Although we can train them to convergence by lowering the learning rate, the obtained results seem underestimated and unreliable. So, we decided to not report the results of CTC and self-conditioned CTC on AISHELL-2. Surprisingly, our proposed UMA method can be successfully trained with a normal learning rate, e.g. 0.0005 in this experiment, which is possibly due to its fewer Conformer layers and shorter output length.
Most of the training parameters for the proposed method are set as the default settings of ESPnet recipes. Warmup steps (from 30k to 40k) and learning rate (from 0.0001 to 0.001) were tuned for each model. The number of training epochs was set within the range of 50 to 70 according to convergence. We don’t use any external language model in decoding. The real-time factor (RTF) is measured on a single Intel(R) Core(TM) i7-11700 CPU @ 2.50GHz. 

\textbf{Integration with self-conditioned CTC.}
It is shown in \cite{higuchi2021comparative} that self-conditioned CTC is one of the best NAR models. The self-conditioned layers help to alleviate the conditional independence assumption of CTC, which can be integrated into the proposed model as well.
 We choose the 6th, 9th, and 12th layers in the encoder, and the 2nd and 4th layers in the decoder as self-conditioned intermediate layers. The training loss weight for the final layer and each intermediate layer are set to 0.5 and 0.1, respectively. 

\subsection{Results and Analysis} 

\textbf{Example of UMA weights.} Fig.~\ref{fig:example} shows an example of UMA weights. It can be seen that the weights clearly exhibit an unimodal characteristic according to text tokens. The unimodal weights are associated with the output hidden units of the encoder. The output of Transformer encoder is well aligned with its input spectrogram, and thus the UMA weights can be used to segment the words in spectrogram. However, the Conformer encoder brings some time shifts relative to the input spectrogram, and the UMA weights are not suitable for word segmentation. The UMA weights of Conformer are more discriminative than the ones of Transformer, in the sense that the weight valleys of Conformer are deeper and clearer than the ones of Transformer. This is possibly because the convolutional layers in Conformer help to learn better speech representations and clearer boundaries among speech units. 

\begin{table}[t]
\vspace{-0.2cm}
\centering
\caption{CER (\%), RTF, and model size on AISHELL-1.}
\label{tab:aishell-1}
\setlength{\tabcolsep}{1.0mm}{
\begin{tabular}{clcccc}
\hline
\multicolumn{2}{c}{\textbf{Model}} & \textbf{dev} & \textbf{test} & \textbf{RTF} & \textbf{Params (M)} \\ 
\hline
\multirow{2}{*}{\rotatebox{90}{AR}}  
& Hybrid (Conformer) & 5.0 & 5.6 & 0.125 & 46.3 \\
& \ + beam search & \textbf{4.3} & \textbf{4.6} & 0.461  & 46.3 \\ 
\hdashline
\multirow{6}{*}{\rotatebox{90}{NAR}} 
& LASO-large \cite{bai2021fast} & 5.9 & 6.6 & - & 80.0 \\
& Paraformer \cite{Gao2022ParaformerFA} & 4.6 & 5.2 & - & - \\
& CTC & 5.6 & 6.1 & 0.052 & 50.4 \\
& Self-conditioned CTC & 4.6 & 4.9 & 0.059 & 51.5 \\
& UMA (prop.) & 4.5 & 4.8 & 0.039 & 42.6 \\
& \ + self-condition & \textbf{4.4} & \textbf{4.7} & 0.045 & 44.7 \\ 
\hline
\end{tabular}
}
\vspace{-0.2cm}
\end{table}

\begin{table}[t]
\vspace{-0.2cm}
\centering
\caption{CER (\%), RTF, and model size on AISHELL-2.}
\label{tab:aishell-2}
\setlength{\tabcolsep}{0.5mm}{
\begin{tabular}{clccccc}
\hline
\multicolumn{2}{c}{\textbf{Model}} & \textbf{android} & \textbf{ios} & \textbf{mic} & \textbf{RTF} & \textbf{Params(M)} \\ 
\hline
\multirow{2}{*}{\rotatebox{90}{AR}}  
& Hybrid (Conformer) & 6.8 & 6.3 & 6.8 & 0.205 & 116.4 \\
& +beam search & \textbf{6.1} & \textbf{5.7} & \textbf{6.1} & 0.954 & 116.4 \\ 
\hdashline
\multirow{4}{*}{\rotatebox{90}{NAR}} 
& LASO-large \cite{bai2021fast} & 7.4  & 6.7 & 7.4 & - & 80.0  \\
& CIF + SAN \cite{dong2020cif} & 6.2  & 5.8 & 6.3 & - & -  \\
& UMA (prop.) & \textbf{6.0} & \textbf{5.3} & 6.0 & 0.085 & 105.4 \\
& + self-condition & \textbf{6.0} & \textbf{5.3} & \textbf{5.9} & 0.098 & 110.4 \\ 
\hline
\end{tabular}
}
\vspace{-0.3cm}
\end{table}

\noindent\textbf{ASR performance of UMA: } Table \ref{tab:hkust}, \ref{tab:aishell-1} and \ref{tab:aishell-2} show the results on HKUST, AISHELL-1 and AISHELL-2, respectively. For most cases, the proposed UMA model outperforms all comparison NAR models. From Table~\ref{tab:hkust}, we can see that the superiority of UMA mainly lies in its much lower substitution error, which demonstrates that explicitly aggregating frames with UMA weights can better learn word representation than implicitly aggregating information by CTC. However, sometimes UMA may lead to some extra deletion errors possibly due to the erroneous merge of multiple tokens, for example in the Transformer and E-Branchformer results of HKUST. Compared to the Transformer encoder, the convolutional-augmented encoders, i.e. Conformer and E-Branchformer, can improve the quality of UMA weights (as shown in Fig.~\ref{fig:example}), and thus promote the advantage of the proposed UMA mechanism.

When integrating self-conditioned layers into the proposed network, the CERs can be further reduced, especially on the HKUST dataset. HKUST is more noisy and spontaneous speaking than AISHELL-1 and AISHELL-2, and the self-conditioned layers help to improve the performance by leveraging the word dependencies. Moreover, the self-conditioned layers in the encoder also help to clarify the word boundaries by aligning the feature sequence and token sequence at intermediate layers, and thus improve the accuracy of UMA weights. This can be verified by the noticeable reduction of deletion errors in the Transformer and E-Branchformer results of HKUST.  
Overall, after integrating self-conditioned layers, the proposed method always achieves comparable ASR performance even with the hybrid CTC/attention + beam search method.  


\noindent \textbf{Model size and RTF analysis: } 
After UMA, the sequence length is reduced from frame level to token level (about one-fifth of the frame-level length), which has two effects on our decoder: i) the convolutional layers are not needed anymore, as the convolutional layers serve for the smoothing of frame-level features, and thus the model size is reduced; ii) the computational complexity is reduced when processing shorter sequence. Table \ref{tab:aishell-1} shows that both the model size and RTF of the proposed model are smaller than the ones of CTC. 

\section{Conclusions}
\label{sec:conclusion}

We propose UMA, a simple yet effective method for NAR ASR, which automatically segments and integrates feature frames for text tokens. Compared to CTC, by integrating feature frames, UMA learns better feature representation and reduces the substitution error, and it reduces the computational complexity as well. By integrating self-conditioned layers, its performance has been further improved. Currently, experiments are only conducted in Mandarin Chinese, and extending UMA to other languages is left for future research.

\vfill\pagebreak

\bibliographystyle{IEEEbib}
\bibliography{refs}

\end{document}